\documentclass{article}

\usepackage{microtype}
\usepackage{graphicx}
\usepackage{subfigure}
\usepackage{booktabs} 
\usepackage{amsmath}
\usepackage{amssymb}
\usepackage{makecell}
\usepackage{multirow}

\usepackage{hyperref}

\usepackage[accepted]{icml2018}

\icmltitlerunning{Convolutional Neural Network Quantization using Generalized Gamma Distribution}

\begin{document}

\twocolumn[
\icmltitle{Convolutional Neural Network Quantization using Generalized Gamma Distribution}




\begin{icmlauthorlist}
\icmlauthor{Doyun Kim}{samsung}
\icmlauthor{Han Young Yim}{samsung}
\icmlauthor{Sanghyuck Ha}{samsung}
\icmlauthor{Changgwun Lee}{samsung}
\icmlauthor{Inyup Kang}{samsung}
\end{icmlauthorlist}

\icmlaffiliation{samsung}{System LSI Business, Samsung Electronics, Gyeonggi-do, South Korea}

\icmlcorrespondingauthor{dyun.kim}{@samsung.com}


\vskip 0.3in
]



\printAffiliationsAndNotice{}  

\begin{abstract}
As edge applications using convolutional neural networks (CNN) models grow, it is becoming necessary to introduce dedicated hardware accelerators in which network parameters and feature-map data are represented with limited precision. In this paper we propose a novel quantization algorithm for energy-efficient deployment of the hardware accelerators. For weights and biases, the optimal bit length of the fractional part is determined so that the quantization error is minimized over their distribution. For feature-map data, meanwhile, their sample distribution is well approximated with the generalized gamma distribution (GGD), and accordingly the optimal quantization step size can be obtained through the asymptotical closed form solution of GGD. The proposed quantization algorithm has a higher signal-to-quantization-noise ratio (SQNR) than other quantization schemes previously proposed for CNNs, and even can be more improved by tuning the quantization parameters, resulting in efficient implementation of the hardware accelerators for CNNs in terms of power consumption and memory bandwidth.
\end{abstract}

\section{Introduction}
\label{submission}

Recent achievements in image processing tasks such as image recognition, object detection, and scene segmentation have been coupled with the application of deep convolutional networks \cite{ref01:szegedy2015going, ref02:ren2015faster, ref03:long2015fully}. As the need for more complex networks increases, we get faced with several implementation issues, i.e. real time processing, limited power budget, and memory bandwidth. For the issues to get resolved, various approaches have been investigated; low-precision \cite{ref04:courbariaux2014training,ref05:courbariaux2015binaryconnect,ref06:hubara2016quantized,ref08:gupta2015deep,ref10:gysel2016hardware,ref11:judd2015reduced,ref14:lin2016fixed}, network compression \cite{ref15:han2015deep}, \cite{ref16:han2016dsd}, small network design \cite{ref17:iandola2016squeezenet}, \cite{ref18:howard2017mobilenets}, and so on. In this paper, meanwhile, we focus on low precision for a power-efficient implementation of hardware accelerators \cite{ref27:horowitz20141}.

Related works are divided into two categories depending on whether the low precision method supports training or not. Firstly, lots of low precision works including training, have been investigated. Through \cite{ref04:courbariaux2014training}, it has been observed that the precision of the multiplication affects the final error after training, and very low precision is sufficient enough not only for inference but also for training. A method to train a network with binary weights is introduced in \cite{ref05:courbariaux2015binaryconnect} through the forward and backward propagations. Moreover, the authors of \cite{ref06:hubara2016quantized} have investigated a method to train quantized neural networks (QNNs) with different bit-widths between inference and training. \cite{ref08:gupta2015deep} deals with a rounding scheme called stochastic rounding, which plays a crucial role in determining network behavior during training. But those works are focused on training under a lower precision, and do not directly support quantization of pre-trained models without re-training.

Next, the other groups have studied low precision schemes only for inference. In \cite{ref10:gysel2016hardware}, a fast and automated framework named Ristretto has been presented for quantization of convolutional neural networks (CNN). The algorithm finds the maximum values of weights and activations to determine the fractional length. In \cite{ref11:judd2015reduced}, the authors have shown how the reduced precision affects the accuracy during inference, and proposed a method to find a low precision configuration, maintaining high accuracy. An optimization problem has been formulated and solved in \cite{ref14:lin2016fixed} for fixed-point bit-width allocation across layers. Also there has been proposed a quantization scheme for inference by optimal uniform quantizer in well-known distributions such as Uniform, Gaussian, Laplace, and gamma.

\begin{figure}[t]
\begin{center}
\includegraphics[width=0.9\linewidth]{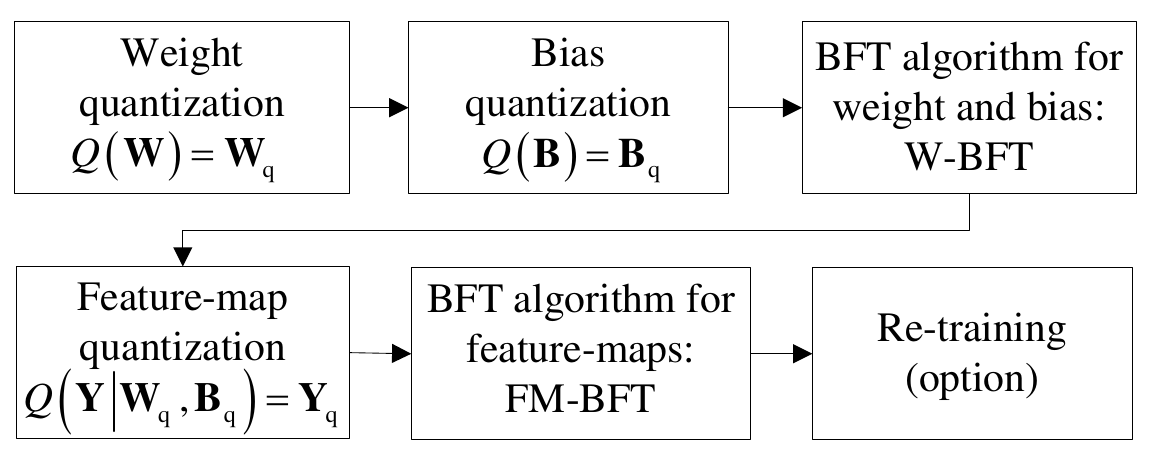}
\end{center}
   \caption{A proposed procedure for network quantization.}
\label{fig:0001}
\end{figure}

In this paper, we propose a novel quantization algorithm to obtain low-precision representations of weights, biases, and feature-maps for CNNs. The algorithm applies different quantization methods to weights/biases and feature-maps, respectively, since the former has deterministic values while the latter can be assumed to have random values. In the algorithm, the quantization of weights and biases precedes the quantization of feature-maps, and re-training the network is considered just optional, as shown in Figure \ref{fig:0001}. For weights and biases, the optimal fractional length is determined in order for the quantization error to be minimized over their distribution. For feature-maps, on the other hand, the generalized gamma distribution (GGD) is used to approximate their sample distribution in each layer. From the asymptotic closed-form equation of GGD \cite{ref19:hui2001asymptotic}, the optimal step size is obtained in terms of signal-to-quantization-noise ratio (SQNR). To further enhance fixed-point performance, we can tune the obtained quantization parameters by an algorithm which we name backward-forward tuning (BFT). Adopting the BFT algorithm on top of the GGD-based quantization, we can effectively deal with the critical issues arising from the discrepancy between SQNR maximization and fixed-point performance optimization in neural networks.

\section{Proposed quantization algorithm}
For both quantization methods for weights/biases and feature-maps in common, we firstly try to determine quantization parameters so that quantization error is minimized. As is already known, however, minimizing quantization error does not necessarily give the optimal quantized network for the classification accuracy. To overcome such a limitation, we introduce an additional step to tune the quantization parameters, following the step to do quantization for minimum quantization error. Note that the tuning step is distinguished from re-training since it does not require training the network.


\subsection{Quantization for SQNR maximization}

\subsubsection{Quantization of weights and biases}

\begin{algorithm}[b]
   \caption{Weight Quantization for $i^{th}$ layer}
   \label{alg:algorithm01}
\begin{algorithmic}
   \STATE {\bfseries Input:} Weights $\boldsymbol{W}_i$, $FL$ set $\boldsymbol{S}_{FL_{\boldsymbol{W}_i}}$
   \FOR{$k=1$ {\bfseries to} $K_{w}$}
   \STATE Initialize $D = 0$:
   \FOR{$j=1$ {\bfseries to} $len(\boldsymbol{W}_i)$}
   \STATE $D~ += \left| w_{i,j} - Q(w_{i,j}, FL)\right|^2$
   \ENDFOR
   \IF {$k = 1$}
   \STATE $D_{min} = D$, $FL_{min} = FL$
   \ELSIF {$D < D_{min}$}
   \STATE $D_{min} = D$, $FL_{min} = FL$
   \ENDIF
   \ENDFOR
   \STATE Determine $FL_{\boldsymbol{W}_i} = FL_{min}$
\end{algorithmic}
\end{algorithm}

Weights and biases should be quantized independently over a pre-trained model of floating-point values, since those have different dynamic ranges each other. The proposed quantization method for weights ($\boldsymbol{W}_i$) for $i^{th}$ layer is shown in Algorithm \ref{alg:algorithm01}. $\boldsymbol{S}_{FL_{\boldsymbol{W}_i}}$ denotes the set including candidates of $FL$ for $i^{th}$ layer as (\ref{eq:algorithm01:FLset}):
\noindent
\begin{equation}
\label{eq:algorithm01:FLset}
\boldsymbol{S}_{FL_{\boldsymbol{W}_i}}=\{ m_{i}, ~m_{i}+1, \cdots, ~m_{i}+K_{w}-1 \},
\end{equation}
\noindent
\noindent where $m_{i} = bw_i - 1-\left \lceil{\log_2\max(|\boldsymbol{W}_i|)}\right \rceil $, $\left \lceil{x}\right \rceil$ is the least integer that is greater than or equal to x, $bw_i$ means a bit-width for the $i^{th}$ layer, minus one is to remove a sign bit, and $K_w$ is the length of $\boldsymbol{S}_{FL_{\boldsymbol{W}_i}}$. In addition, $len(\boldsymbol{v})$ is a length of a vector $\boldsymbol{v}$, $D$ is a quantization error for $k^{th}$ $FL$ value, $FL_{min}$ is the $FL$ value with the minimum quantization error $D_{min}$, and $Q(w_{i,j}, FL)$ is quantized weight of $w_{i,j}$ by $FL$. Finally, Algorithm \ref{alg:algorithm01} is to determine $FL_{\boldsymbol{W}_i}$ to minimize overall quantization error as follows: $FL_{\boldsymbol{W}_i}=\arg \min_{FL} D(FL)$.
For bias quantization, all you have to do is to replace $\boldsymbol{W}_i$ with $\boldsymbol{b}_i$. As described in Algorithm \ref{alg:algorithm01}, a simple method is used for weight/bias quantization in order to minimize the quantization error separately for each layer. It is sufficient to consider $K_w$ = 2 in $\boldsymbol{S}_{FL_{\boldsymbol{W}_i}}$, since an overload distortion rapidly increases as a $FL$ increases. 

\subsubsection{Quantization of feature-maps}
\begin{figure}[tb]
\begin{center}
\includegraphics[width=0.9\linewidth]{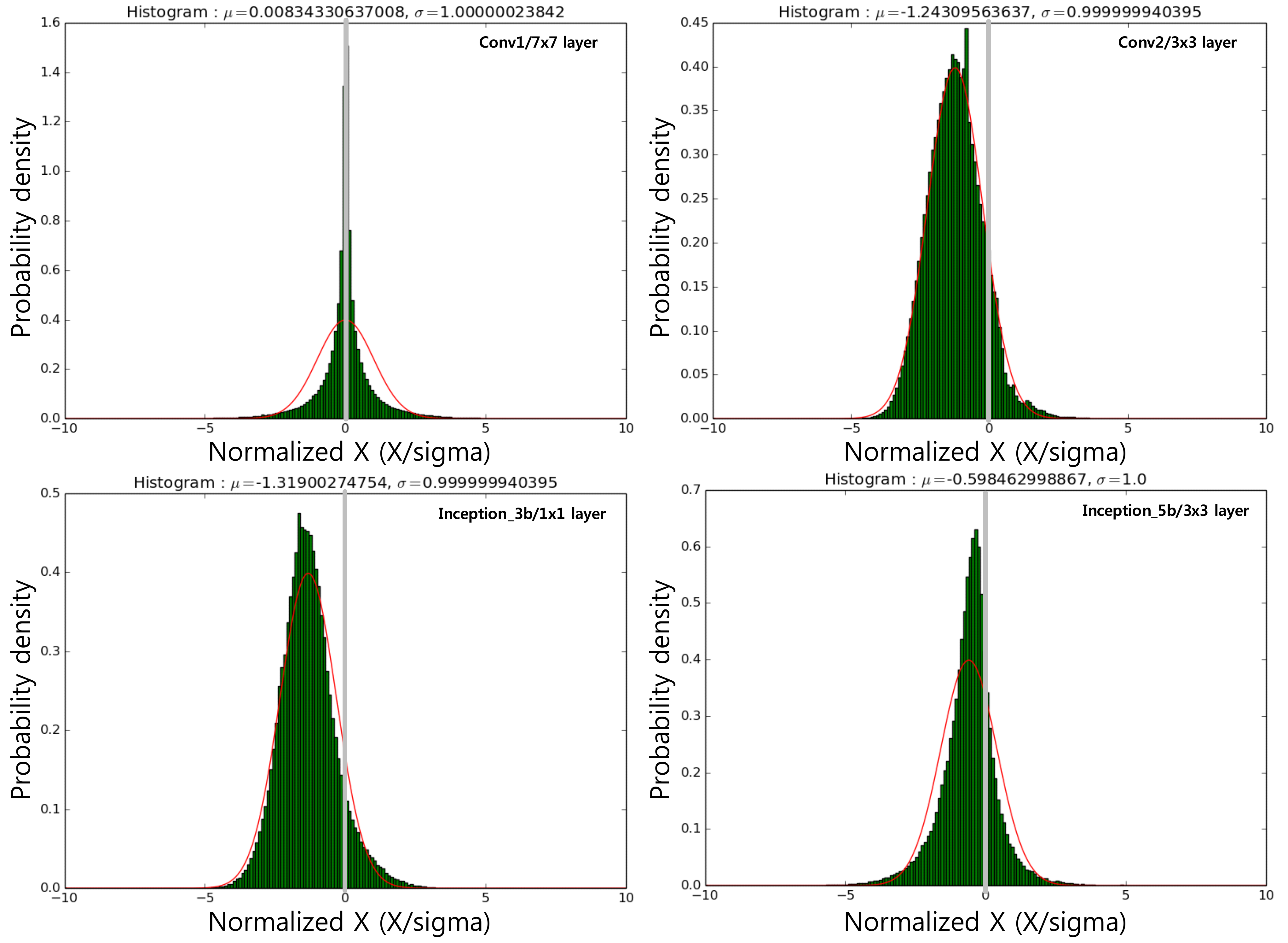}
\end{center}
   \caption{Comparisons between distribution of pre-activations (green colored histogram) and Gaussian distribution (red line) for four convolutional layers.}
\label{fig:0002}
\end{figure}

To give a rough view of the output feature-map distribution, take a look at GoogLeNet \cite{ref01:szegedy2015going} as an example. GoogLeNet consists of 57 convolutional layers with Rectified Linear Unit (ReLU) and a fully connected layer in inference. Figure \ref{fig:0002} includes four graphs, each of which is derived from a convolutional layer of GoogLeNet and represents two distributions on the same graph: 1) the sample distribution of pre-activation values and 2) the Gaussian distribution with unit variance. In the upper left graph, which shows the distribution of pre-activation values in the first convolutional layer of GoogLeNet, $conv1/7\times7$-layer, it is observed that the distribution has a near zero mean and an almost symmetric density shape since subtracting the mean from data is included in the data layer. Except the $conv1/7\times7$-layer, meanwhile, the others have a non-zero mean and are even not symmetric about the non-zero mean.
The most conventional way for feature-map quantization is to approximate the sample distribution of pre-activations with a symmetric distribution like the Gaussian distribution or the Laplace distribution. Unfortunately, however, such a quantization method is sometimes not optimal in terms of SQNR observed at the input of the next layer for the following reasons: 1) The mismatch between asymmetric sample distributions and symmetric model distributions causes critical overload distortion, resulting in difficulty finding the optimal step size in quantization. 2) When assuming that the mean of pre-activations is none-zero and ReLU is used as an activation function, the efficient quantization is to consider only the pre-activations of a positive value since all the pre-activations of a negative value is supposed to be mapped to zero after passing through ReLU activation.

\begin{figure}[tb]
\begin{center}
\includegraphics[width=0.9\linewidth]{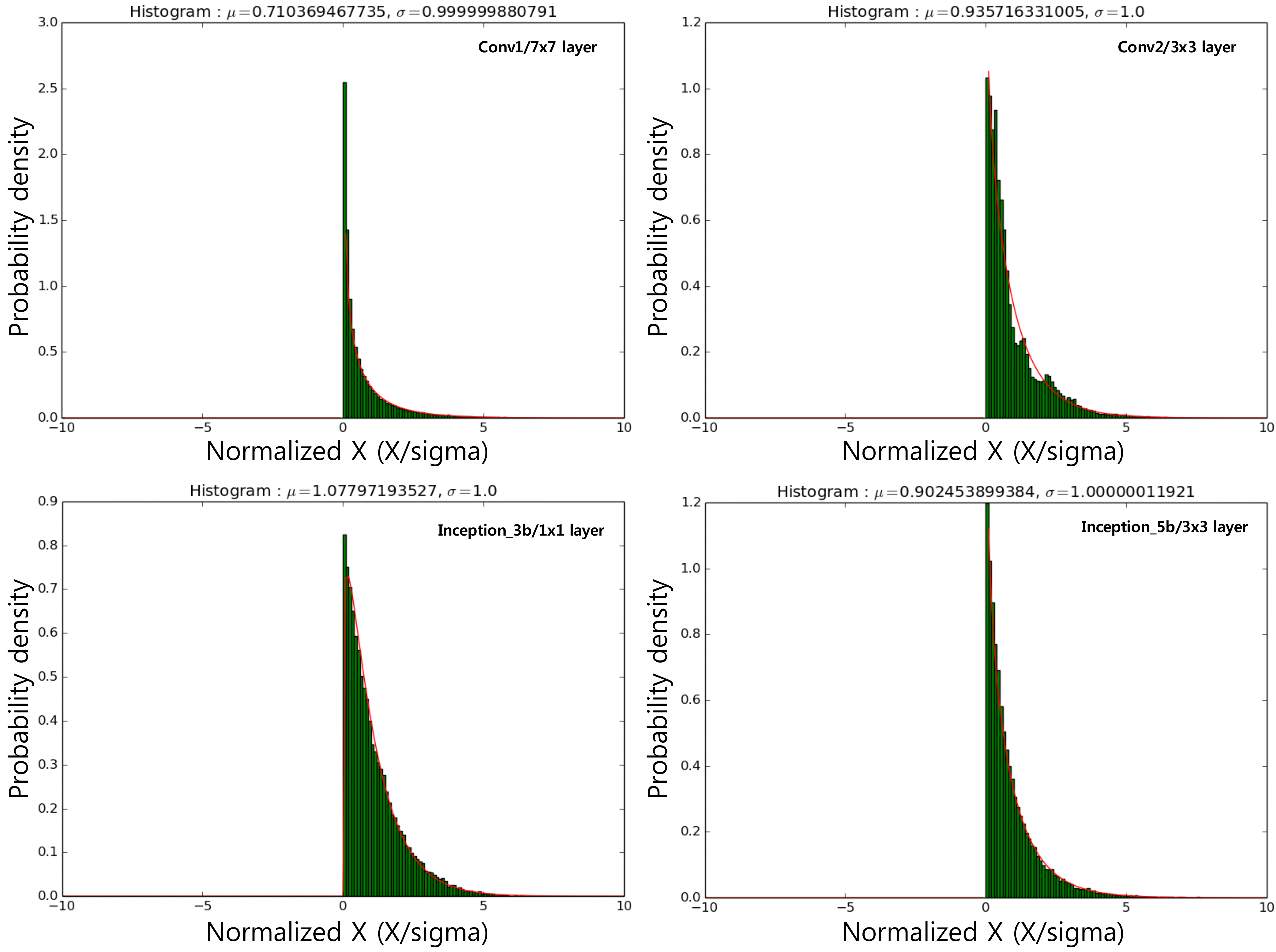}
\end{center}
   \caption{Comparisons between distribution of activation values (green colored histogram) and gamma distribution (red line) for four convolutional layers when all zeroes are excluded in the evaluations and in the plots.}
\label{fig:0004}
\end{figure}

%

To overcome the problems of the feature-map quantization based on a symmetric distribution, we tried to search for an alternative approach immune to non-zero mean and asymmetric distribution. We focused on the distribution of activations instead of pre-activations. Assuming ReLU is used as an activation function, the distribution of activations of a positive value is single-sided. For quantization, we don't need to include the activations of zero value in the distribution. The resultant sample distribution has a good match in the positive part for a symmetric distribution such as the Gaussian distribution and the Laplace distribution, and a better match for a distribution with heavy tails such as the gamma distribution, as shown in Figure \ref{fig:0004}.

\textbf{Generalized gamma distribution:} The probability density function of the gamma distribution is represented as follows:
\begin{equation}
\label{eq:Gamma dist}
\frac{1}{\Gamma(\kappa)\theta^{\kappa}}x^{\kappa - 1}e^{-\frac{x}{\theta}}
\end{equation}
\noindent where $\Gamma (\tau)=\int_{0}^{\infty}x^{\tau - 1}e^{-x}dx$ (gamma function), $\kappa$ and $\theta$ denote a shape and a scale parameters in the gamma distribution, respectively. And there are simple relationships between mean, variance, $\kappa$ and $\theta$ as
\begin{equation}
\label{eq:Mean and Var est}
E[X] = \kappa\theta~ \textnormal{and} ~Var[X] = \kappa\theta^2
\end{equation}
We calculated $\kappa$ and $\theta$ with sample mean and variance, while all zeroes being excluded. Then, we plot the sample distribution and the gamma distribution with calculated $\kappa$ and $\theta$ in Figure \ref{fig:0004}. Through the results, the gamma distribution and sample distribution are found well-matched, especially for the tails of those. Therefore, if the optimal step size is found based on the gamma distribution parameterized through our calculation, we have an optimal solution to the feature-map quantization in terms of SQNR.

In \cite{ref19:hui2001asymptotic}, minimum mean-square error (MSE)-optimal N-level uniform scalar quantizers are designed numerically and their asymptotic characteristics tabulated in a closed form. The distribution of a GGD is represented by (\ref{eq:General Gamma dist}):
\begin{equation}
\label{eq:General Gamma dist}
p(x)=\mu |x|^{\beta}e^{-\lambda|x|^{\alpha}}
\end{equation}
\noindent where $\alpha>0,~\beta>-1$ is the exponential decay parameter, $\mu$ and $\lambda$ are in \cite{ref19:hui2001asymptotic}. When $\alpha$=1, such densities reduce to the gamma density. The asymptotic equations of the support length $\hat{L}_N$ and the overall distortion $\hat{D}_N$ have been derived as follows:
\begin{equation}
\label{eq:support length}
\hat{L}_N = \left[
\begin{aligned}  & \frac{2\ln N}{\lambda} - \left( 2-\frac{1+\beta}{\alpha}\right) \frac{\ln\ln N}{\lambda} \\
& - \frac{1}{\lambda} \ln \left( \frac{2^{1-(1+\beta)/\alpha} \alpha^2 \lambda^{(1+\beta)/\alpha}}{3\mu} \right) + \epsilon_N
\end{aligned}
\right] ^{1/\alpha}
\end{equation}
\noindent where $\epsilon_N$ is the correction term of closed form as (\ref{eq:epsilon_gamma}) \cite{ref19:hui2001asymptotic}. $\hat{L}_N$ closely approximates the actual $L_N$ even at N = 4 or 8. And by using a $\hat{L}_N$, we also calculate the overall distortion $\hat{D}_N(\hat{L}_N)$ in (\ref{eq:overall distortion}).
\begin{equation}
\label{eq:epsilon_gamma}
\begin{aligned}
&\epsilon_N = \frac{1}{\lambda}\ln \cdot \\
&\left[
\begin{aligned}
&\left( 1+\frac{2\alpha\ln N}{N} \right) \left( 1+\frac{3 - 3\alpha + 2\beta}{2\alpha\ln N} \right) \cdot\\
&\left( 1+\frac{1}{2\ln N} \left(
\begin{aligned}
&\left( 2-\frac{1+\beta}{\alpha} \right) \ln \ln N + \\
&\ln \left( \Phi \right)
\end{aligned}
 \right) \right)^{2-(1+\beta)/\alpha}
\end{aligned}
\right]
\end{aligned}
\end{equation}
\noindent where $\Phi$ = $\frac{2^{1-(1+\beta)/\alpha} \alpha^2 \lambda^{(1+\beta)/\alpha}}{3\mu}$.
\begin{equation}
\label{eq:overall distortion}
\hat{D}_N(\hat{L}_N) = \frac{1}{12} \left( \frac{2\hat{L}_N}{N}\right)^2 + \frac{4\mu}{(\alpha\lambda)^3}\frac{e^{-\lambda \hat{L}_{N}^{\alpha}}}{\hat{L}_{N}^{3\alpha-\beta-3}}
\end{equation}
Finally, these equations for optimal step size and distortion extend to densities with one-sided infinite support such as single-sided gamma of (\ref{eq:Gamma dist}). Specifically, the optimal N-level uniform scalar quantizer with support interval $[0,~N\Delta_N]$ for a density with support $[0,~\infty]$ has step size, $\Delta_N=\Delta_{2N}$, and distortion, $D_N=D_{2N}$, where $\Delta_{2N}$ and $D_{2N}$ are the step size and distortion for the optimal $2N$-level symmetric quantizer for the symmetric density, $\tilde{p}(x)=(p(x)+p(-x)) /2$.

\textbf{Single-sided distribution:} By using the GGD mentioned in previous section, we propose a quantization method for the activations with single-sided sample distribution in Algorithm \ref{alg:algorithm02}. Here, $m_x$ and $\sigma^2_x$ are mean and variance to be estimated from samples after activation function of $i^{th}$ layer, where zeroes have to be excluded. Additionally, $\Gamma(~)$ is the gamma function, $\lfloor{v}\rfloor$ is the greatest integer that is less than or equal to $v$, $x_{i,j}$ is the $j^{th}$ sample of $i^{th}$ layer, $\boldsymbol{x}_i$ is the sample set of $i^{th}$ layer, and $Q(x_{i,j}, FL)$ means the quantized value of $x_{i,j}$ with fractional length $FL$. We can optionally extend the algorithm to GGD by estimating parameters $(\alpha,~\beta,~\lambda,~\mu)$ with various approaches \cite{ref23:stacy1965parameter}, \cite{ref24:gomes2008parameter}. It is important that zeroes have to be excluded when estimating the stochastic characteristics. We recommend to use the default mode for higher SQNRs and the fast mode for shorter simulation time.

%
%
%
%
%
%
%
%
%

\begin{algorithm}[t]
   \caption{FM Quantization for a single-sided distribution for $i^{th}$ layer}
   \label{alg:algorithm02}
\begin{algorithmic}
   \STATE {\bfseries Input:} Single-sided sample set $\boldsymbol{x}_i$, $m_x$, $\sigma^2_x$, $N$

   \STATE {\bfseries Begin:}
   \STATE Estimate parameters $(\alpha,~\beta,~\lambda,~\mu)$ of a GGD:
   \STATE $\alpha=1,~\beta=m_x^2/\sigma^2_x-1,~\lambda=m_x/\sigma^2_x,~\mu=\frac{\lambda^{m_x^2/\sigma^2_x}}{2\Gamma(m_x^2/\sigma^2_x)}$

   \STATE Calculate $\hat{L}_N$ by (\ref{eq:support length}) and (\ref{eq:epsilon_gamma}) with $(\alpha,~\beta,~\lambda,~\mu)$
   \STATE Calculate the optimal step size, $\hat{\Delta}_N = 2\hat{L}_N /N$

   \STATE Determine the $FL$ set, $\boldsymbol{S}_{FL_{\boldsymbol{x}_i}}$:
   \STATE $\boldsymbol{S}_{FL_{\boldsymbol{x}_i}} = \{ -\lceil{\log_2 \hat{\Delta}_N}\rceil, ~-\lfloor{\log_2 \hat{\Delta}_N}\rfloor\} $.

   \FOR{k=1 {\bfseries to} $len(\boldsymbol{S}_{FL_{\boldsymbol{x}_i}})$}

   \STATE $FL= \boldsymbol{S}_{FL_{\boldsymbol{x}_i}}(k)$

   \IF {$\text{mode} = \text{default}$}
   \STATE $\hat{D}_N=\sum_{x_{i,j}\in \boldsymbol{x}_i} \left| x_{i,j} - Q(x_{i,j}, FL)\right|^2$
   \ELSIF {$\text{mode} = \text{fast}$}
   \STATE $\hat{\Delta}_{N} = 2^{-FL}$
   \STATE $\hat{L}_{N} = N\hat{\Delta}_{N}/2$
   \STATE $\hat{D}_{N} = \hat{D}_{N}(\hat{L}_{N})$ in (\ref{eq:overall distortion})
   \ENDIF

   \IF {$k = 1$}
   \STATE $D_{min} = \hat{D}_N$, $FL_{min} = FL$
   \ELSIF {$\hat{D}_N < D_{min}$}
   \STATE $D_{min} = \hat{D}_N$, $FL_{min} = FL$
   \ENDIF

   \ENDFOR
   \STATE Determine $FL_{\boldsymbol{x}_i} = FL_{min}$

\end{algorithmic}
\end{algorithm}

\begin{figure}[]
\begin{center}
\includegraphics[width=0.9\linewidth]{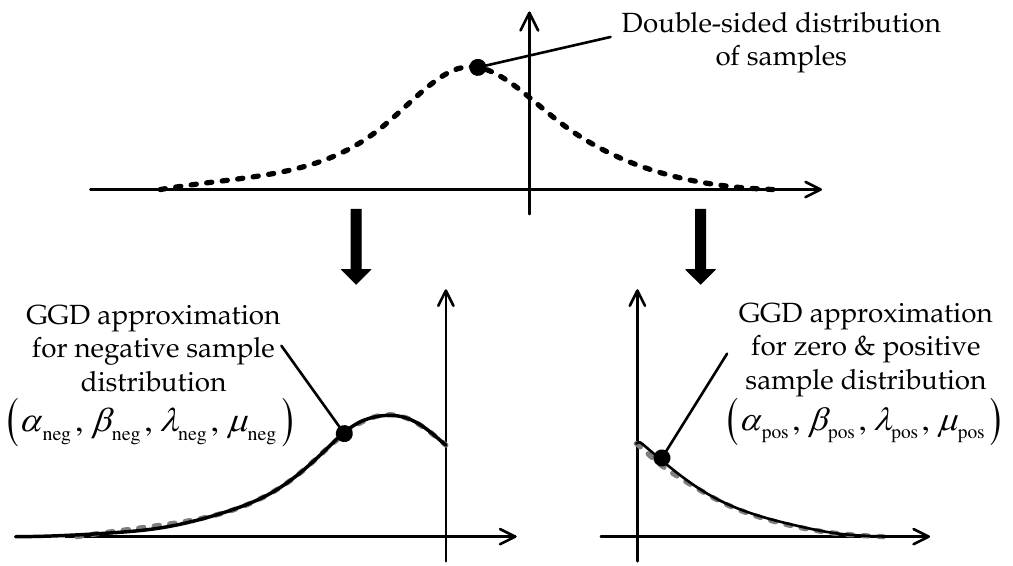}
\end{center}
   \caption{Concept of feature-map quantization for sample set with double-sided distribution.}
\label{fig:0005}
\end{figure}

\textbf{Double-sided distribution:} We also propose a quantization method for the activations with double-sided sample distribution. In CNNs, a double-sided distribution can be generated from pre-activation samples or from the output of sigmoid, hyper-tangent, or p-ReLU activation functions. In GoogLeNet, the loss3/classifier-layer has a double-sided distribution because its output is connected to the soft-max-layer.
To quantize the samples with a double-sided distribution, we first divide the samples into two groups, as shown in Figure \ref{fig:0005}. Next, we estimate parameters $(\alpha,~\beta,~\lambda,~\mu)$ of the GGD for each group and find the optimal fractional length. Algorithm details are described in Algorithm \ref{alg:algorithm03}. Here, $N$ is a number of quantization levels, $\rho=(\text{number of negative samples})/{(\text{number of total samples})}$. 

\begin{algorithm}[]
   \caption{FM Quantization for a double-sided distribution for $i^{th}$ layer}
   \label{alg:algorithm03}
\begin{algorithmic}
   \STATE {\bfseries Input:} Double-sided sample set $\boldsymbol{x}_i$, $N$, $\rho$, empty set $\boldsymbol{C}$

   \STATE {\bfseries Begin:}
   \STATE Divide $\boldsymbol{x}_i$ into two groups: a negative sample group ($neg$) and a zero\&positive sample group ($pos$).

   \FOR {$g$ {\bfseries in} $[neg, ~pos]$ }
   \STATE {\bfseries Input: } $m_x$ and $\sigma^2_x$ for each group
   \STATE Estimate parameters $(\alpha,~\beta,~\lambda,~\mu)$ of a GGD:
   \STATE $\alpha=1,~\beta=m_x^2/\sigma^2_x-1,~\lambda=m_x/\sigma^2_x,~\mu=\frac{\lambda^{m_x^2/\sigma^2_x}}{2\Gamma(m_x^2/\sigma^2_x)}$

   \STATE Calculate $\hat{L}_N$ by (\ref{eq:support length}) and (\ref{eq:epsilon_gamma}) with $(\alpha,~\beta,~\lambda,~\mu)$
   \STATE Calculate the optimal step size, $\hat{\Delta}_N = 2\hat{L}_N /N$
   \STATE  $\boldsymbol{C} \gets [ \boldsymbol{C},~-\lceil{\log_2 \hat{\Delta}_{N}} \rceil,~-\lfloor\log_2 \hat{\Delta}_{N}\rfloor]$
   \STATE $(\alpha_{g},~\beta_{g},~\lambda_{g},~\mu_{g}) \gets (\alpha,~\beta,~\lambda,~\mu)$
   \ENDFOR

   \STATE Determine the $FL$ set, $\boldsymbol{S}_{FL_{\boldsymbol{x}_i}}$:
   \STATE $\boldsymbol{S}_{FL_{\boldsymbol{x}_i}} = \{ x | \min(\boldsymbol{C}) \leq x \leq \max(\boldsymbol{C}) \} $

   \FOR{k=1 {\bfseries to} $len(\boldsymbol{S}_{FL_{\boldsymbol{x}_i}})$}
   \STATE $FL= \boldsymbol{S}_{FL_{\boldsymbol{x}_i}}(k)$
   \IF {$\text{mode} = \text{default}$}
   \STATE $\hat{D}_{D,N}=\sum_{x_{i,j}\in \boldsymbol{x}_i} \left| x_{i,j} - Q(x_{i,j}, FL)\right|^2$
   \ELSIF {$\text{mode} = \text{fast}$}
   \STATE $\hat{\Delta}_{N} = 2^{-FL}$
   \STATE $\hat{L}_{N} = N\hat{\Delta}_{N}/2$
   \STATE Calculate overall distortion $\hat{D}_{D,N}$ by using (\ref{eq:overall distortion}):
   \STATE $\hat{D}_{neg}=\hat{D}_{N}(\hat{L}_{N})$ with $\alpha_{neg},\beta_{neg},\lambda_{neg},\mu_{neg}$\\
   $\hat{D}_{pos}=\hat{D}_{N}(\hat{L}_{N})$ with $\alpha_{pos},\beta_{pos},\lambda_{pos},\mu_{pos}$\\
   \STATE $\hat{D}_{D,N}= \rho\hat{D}_{neg} + (1-\rho)\hat{D}_{pos}$
   \ENDIF

   \IF {$k = 1$}
   \STATE $D_{min} = \hat{D}_{D,N}$, $FL_{min} = FL$
   \ELSIF {$\hat{D}_{D,N} < D_{min}$}
   \STATE $D_{min} = \hat{D}_{D,N}$, $FL_{min} = FL$
   \ENDIF

   \ENDFOR
   \STATE Determine $FL_{\boldsymbol{x}_i} = FL_{min}$

\end{algorithmic}
\end{algorithm}

\subsection{Tuning quantization parameters}
The BFT algorithm tunes the quantization parameters which have already been set for SQNR maximization since the maximization does not always come with the highest performance in quantized networks.
For example, assume that the loss3/classifier layer of GoogLeNet has been quantized resulting in a saturation threshold with which the maximum SQNR is achieved. But, if the ratio of the layer outputs whose value is exceeding the threshold is much greater than 1/1000, Top-1 accuracy will be dropped since a plural output samples of the loss3/classifier layer have the same highest probability.

As explained in Algorithm \ref{alg:algorithm04}, the BFT algorithm finds the fractional length in each layer to maximize the network performance, i.e. Top-1 or Top-5 accuracy for GoogLeNet. Also, the algorithm is applicable to weights, biases, and feature-maps, independently. The reason why the algorithm is performed in a backward direction and then in a forward direction is that the network performance is more affected by the change of the fractional length in the layers close to the output, and accordingly it is efficient to find the best fractional length in a backward direction first, i.e. from the loss layer to the input layer. To reflect the change of layer output distribution due to the updated fractional lengths, it is necessary to find the fractional length again for each layer in a forward direction, i.e. from the input layer to the loss layer. In Algorithm \ref{alg:algorithm04}, $\boldsymbol{S}_{BFT}$ denotes a set of layer indices according to the tuning order and $T$ is the number of performance metrics in a network. In case of GoogLeNet, $T = 2$ since GoogLeNet has two performance metrics, Top-1 accuracy and Top-5 accuracy. $P_n$ and $c_n$ denote the $n^{th}$ performance metric and the weight of the metric, respectively. $K_{BFT}$ is a positive integer number indicating the size of a searching window for tuning, and its default value is 1.

\begin{algorithm}[t]
   \caption{Backward-Forward Tuning (BFT)}
   \label{alg:algorithm04}
\begin{algorithmic}

   \STATE {\bfseries Input:} layer index set $\boldsymbol{S}_{BFT}$, $K_{BFT}$

   \STATE {\bfseries Begin:}
   \STATE Start the BFT procedure:
   \FOR{$l=1$ {\bfseries to} $len(\boldsymbol{S}_{BFT})$}
   \STATE {$i \gets \boldsymbol{S}_{BFT}(l)$}
   \STATE {$FL_i \gets FL_{\boldsymbol{W}_i} (\text{or} ~FL_{\boldsymbol{x}_i} )$}
   \STATE $\boldsymbol{S}_{FL,i}=\{ x|FL_i-K_{BFT} \leq x \leq FL_i+K_{BFT}\}$
   \FOR{$k=1$ {\bfseries to} $len(\boldsymbol{S}_{FL,i})$}
   \STATE Run the network with $\boldsymbol{S}_{FL,i}(k)$
   \STATE Get performance set $\{P_0, \cdots, P_{T-1}\}$
   \STATE Compute $P_{\text{overall}}(k) = \sum_{n=0}^{T-1} c_n P_n$
   \ENDFOR
   \STATE $ m^* = \arg \max_{m} P_{\text{overall}}(m) $
   \STATE $ FL_{\boldsymbol{W}_i} (\text{or} ~FL_{\boldsymbol{x}_i} ) \gets \boldsymbol{S}_{FL,i}(m^*)$
   \ENDFOR

\end{algorithmic}
\end{algorithm}


\section{Performance evaluation}
We evaluate the fixed-point performance of GoogLeNet \cite{ref01:szegedy2015going} with the proposed quantization algorithm on the ImageNet \cite{ref25:deng2009imagenet} (ILSVRC 2012) validation dataset, which has 50,000 images for inference, in the Caffe framework \cite{ref26:jia2014caffe}. The inference step is operated with 57 convolutional layers, one fully connected layer, and two local response normalization (LRN) layers in GoogLeNet to get Top-1 and Top-5 accuracies. We apply the proposed algorithm to two types of layers, convolutional layer and fully connected layer. The parameter $\alpha$ is set to be 1 for feature-map quantization and $K_{BFT}$ to be 1 in the BFT algorithm for low computing complexity. We set $c_0$=1 and $c_1$=0 in the BFT algorithm for focusing on the Top-1 accuracy. We estimate samples’ mean ($m_x$) and variance ($\sigma_x^2$) of all layers using 6400 training images on inference operations with the mini-batch size of 32.

\begin{figure}[t]
\begin{center}
\includegraphics[width=0.9\linewidth]{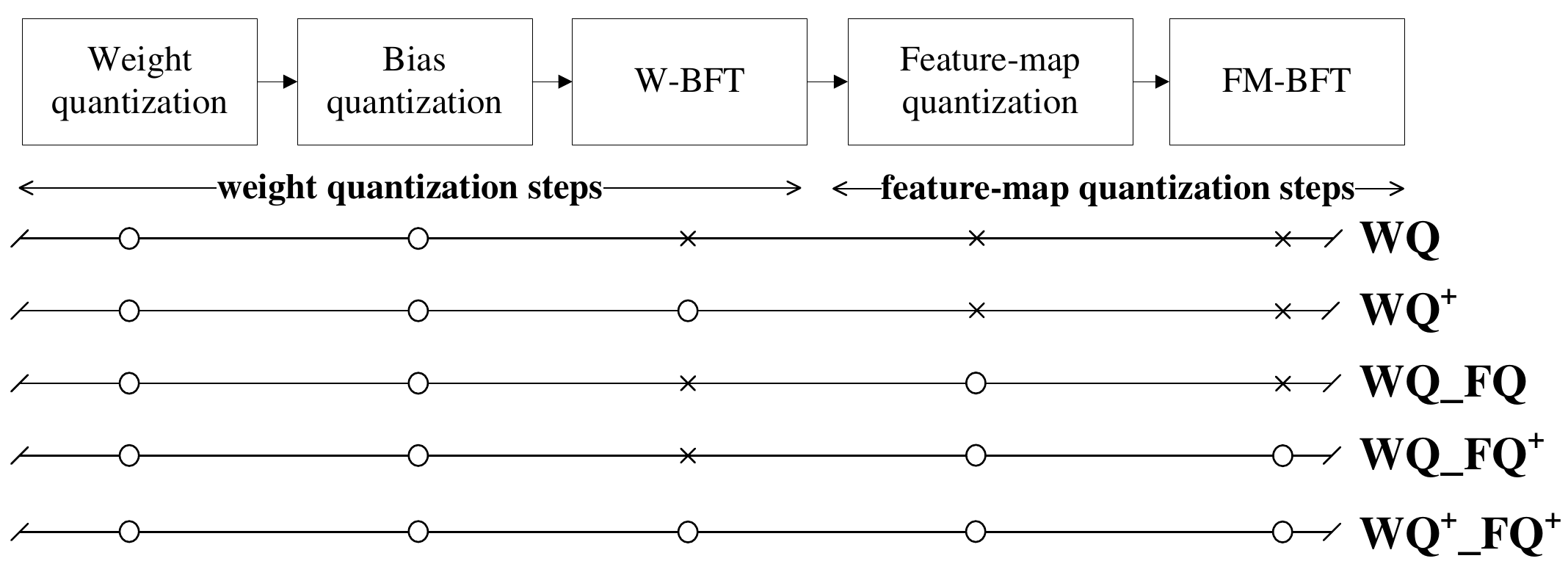}
\end{center}
   \caption{Explanation about five schemes and two quantization steps used in evaluations: A circle and a cross on lines mean whether the corresponding step is operated or not, respectively.}
\label{fig:0008}
\end{figure}

To explain the proposed quantization algorithm clearly, we use five different names, 1) $WQ$, 2) $WQ^+$, 3) $WQ\_FQ$, 4) $WQ\_FQ^+$, and 5) $WQ^+\_FQ^+$, according to the processing steps included for quantization, as shown in Figure \ref{fig:0008}.
$WQ$ includes only weight/bias quantization steps, while feature-map having floating values. $WQ^+$ means $WQ$ followed by W-BFT. $WQ\_FQ$, $WQ\_FQ^+$, and $WQ^+\_FQ^+$ denote three types of whole consecutive procedure including weight and feature-map quantization steps, where $+$ means the use of BFT algorithm at the end of each quantization step.

As a reference we use the Ristretto-based quantization scheme, where the criteria to find fractional lengths for weight/bias and feature-map in the $i^{th}$ layer have been experimentally determined given a bit width $bw$, as shown in (\ref{eq:fl_ristretto}). Note that we use the Ristretto-based approach only for quantization without re-training \cite{ref10:gysel2016hardware}.
\begin{equation}
\label{eq:fl_ristretto}
\begin{aligned}
&fl_{\boldsymbol{W}_i}=bw-1-\lceil{\log_2 \max(|\boldsymbol{W}_i)} \rceil \\
&fl_{\boldsymbol{x}_i}=bw-\lceil{\log_2 \max(\boldsymbol{x}_i)} \rceil \\
\end{aligned}
\end{equation}

\begin{figure}[t]
\begin{center}
\includegraphics[width=0.9\linewidth]{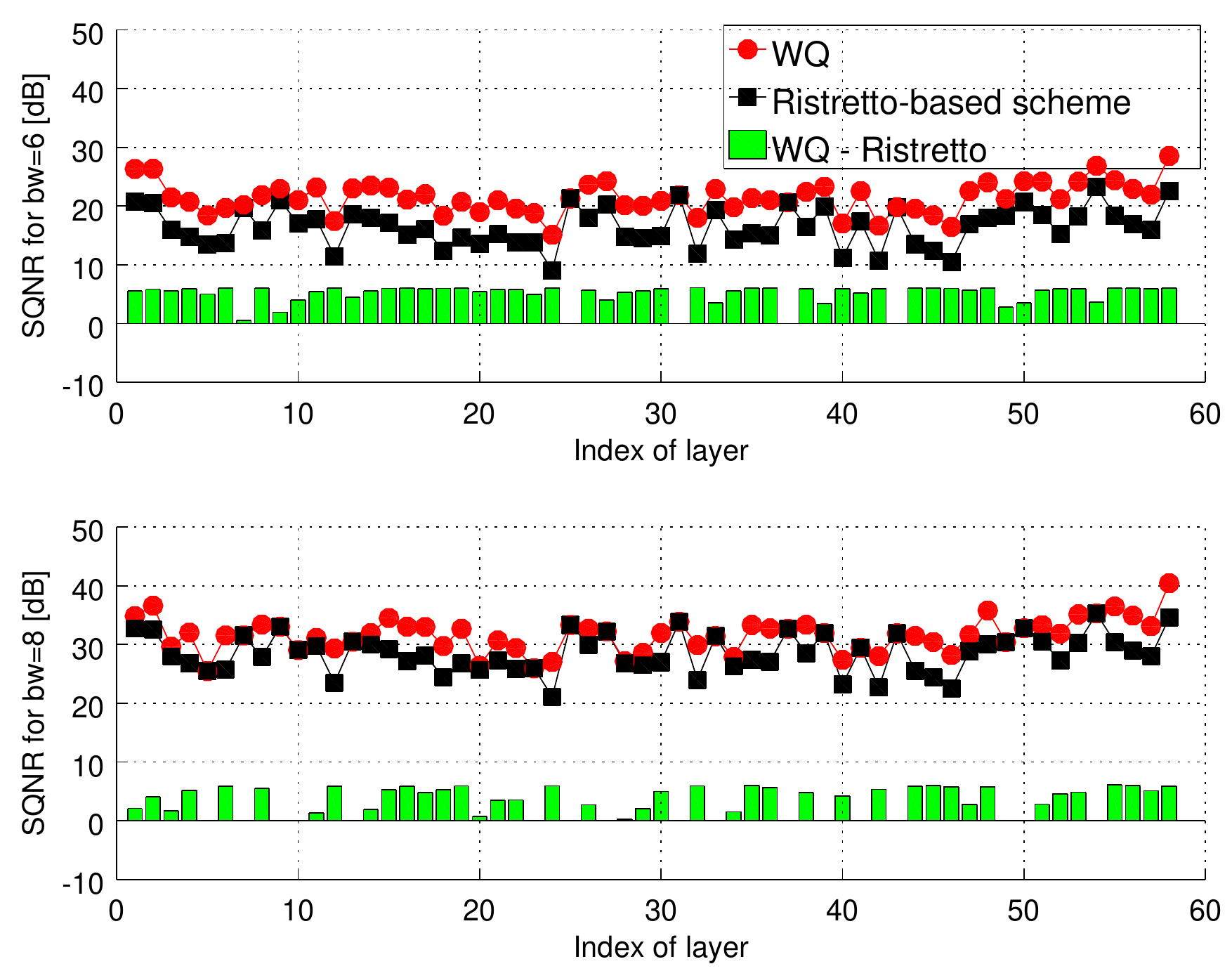}
\end{center}
   \caption{SQNRs for all layers after weight/bias quantization.}
\label{fig:0010}
\end{figure}

\begin{figure}[t]
\begin{center}
\includegraphics[width=0.9\linewidth]{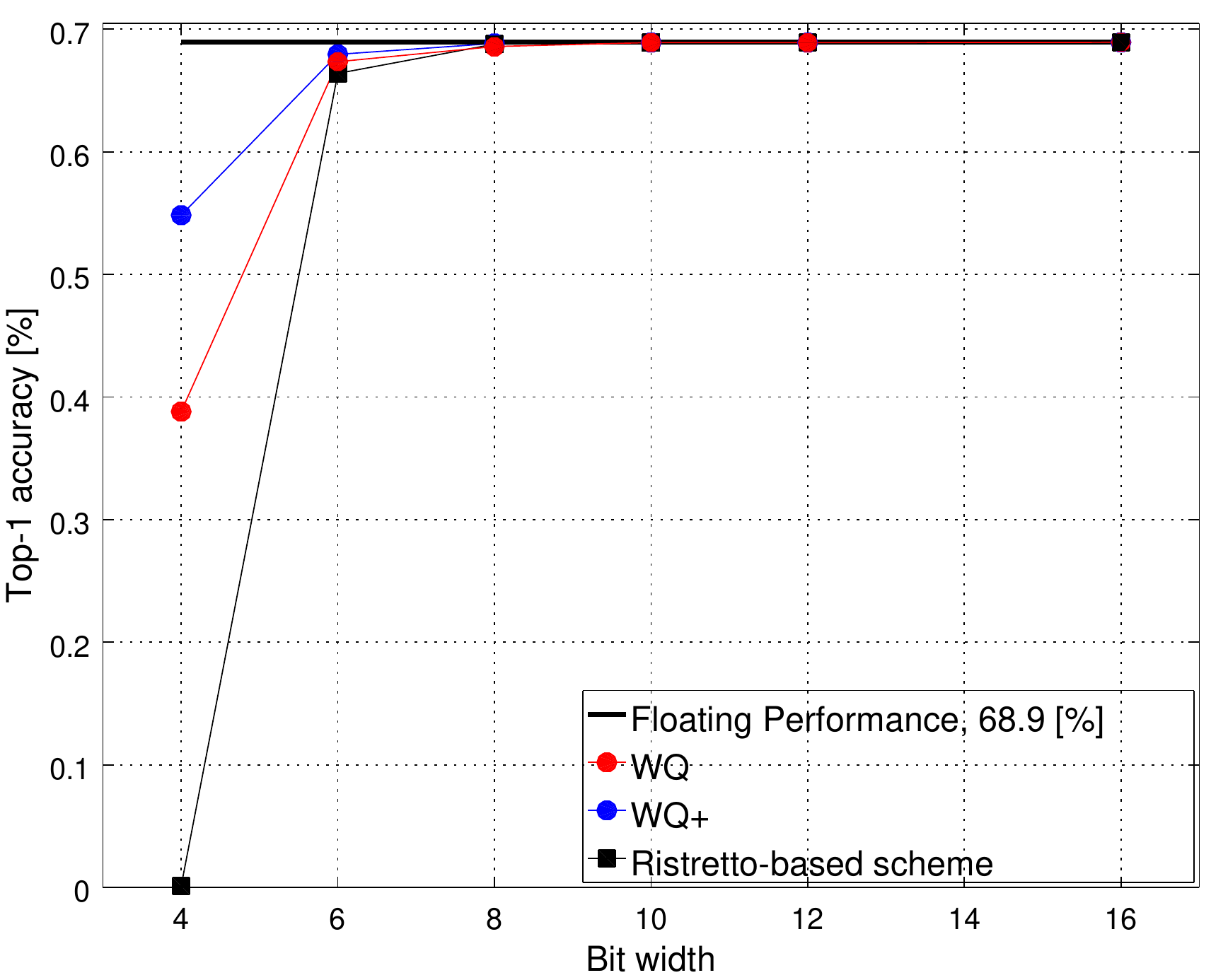}
\end{center}
   \caption{Top-1 accuracy of GoogLeNet after weight/bias quantization.}
\label{fig:0011}
\end{figure}

\subsection{Evaluation of weight quantization steps}
To evaluate the weight/bias quantization scheme, we define weight quantization steps in Figure \ref{fig:0008}. In the steps, we can concentrate on the performance comparison of weight/bias quantization under floating-valued feature-maps. Figure \ref{fig:0010} shows SQNR of each scheme and SQNR difference between them for all layers. It is observed that most layers have SQNR gain through the proposed scheme for both $bw$ = 6 and $bw$ = 8. How much gain can be achieved in terms of Top-1 accuracy owing to better SQNR? The answer is shown in Figure \ref{fig:0011}, where Top-1 accuracy is 68.9 \% in a floating network with pre-trained BVLC GoogLeNet model\footnote{Reference model is from Caffe model zoo}. All three schemes can almost achieve floating-level performance when $bw \geq 8$. When $bw < 8$, however, floating-level accuracy cannot be achieved even with $WQ$ and $WQ^+$, which show lots of performance gain over the Ristretto-based scheme though. Three observations in the results are as follows: 1) $WQ$ is effective to reduce performance loss in a quantized neural network, 2) $WQ^+$ provides an excellent performance gain for a small $bw$, and 3) there may exist a required SQNR level of a weight quantization step to achieve the floating-level performance in a deep neural network. Unfortunately, the relationship between accuracy and SQNR has not yet been clarified.


\subsection{Evaluation of feature-map quantization steps}
\begin{figure}[tb]
\begin{center}
\includegraphics[width=0.9\linewidth]{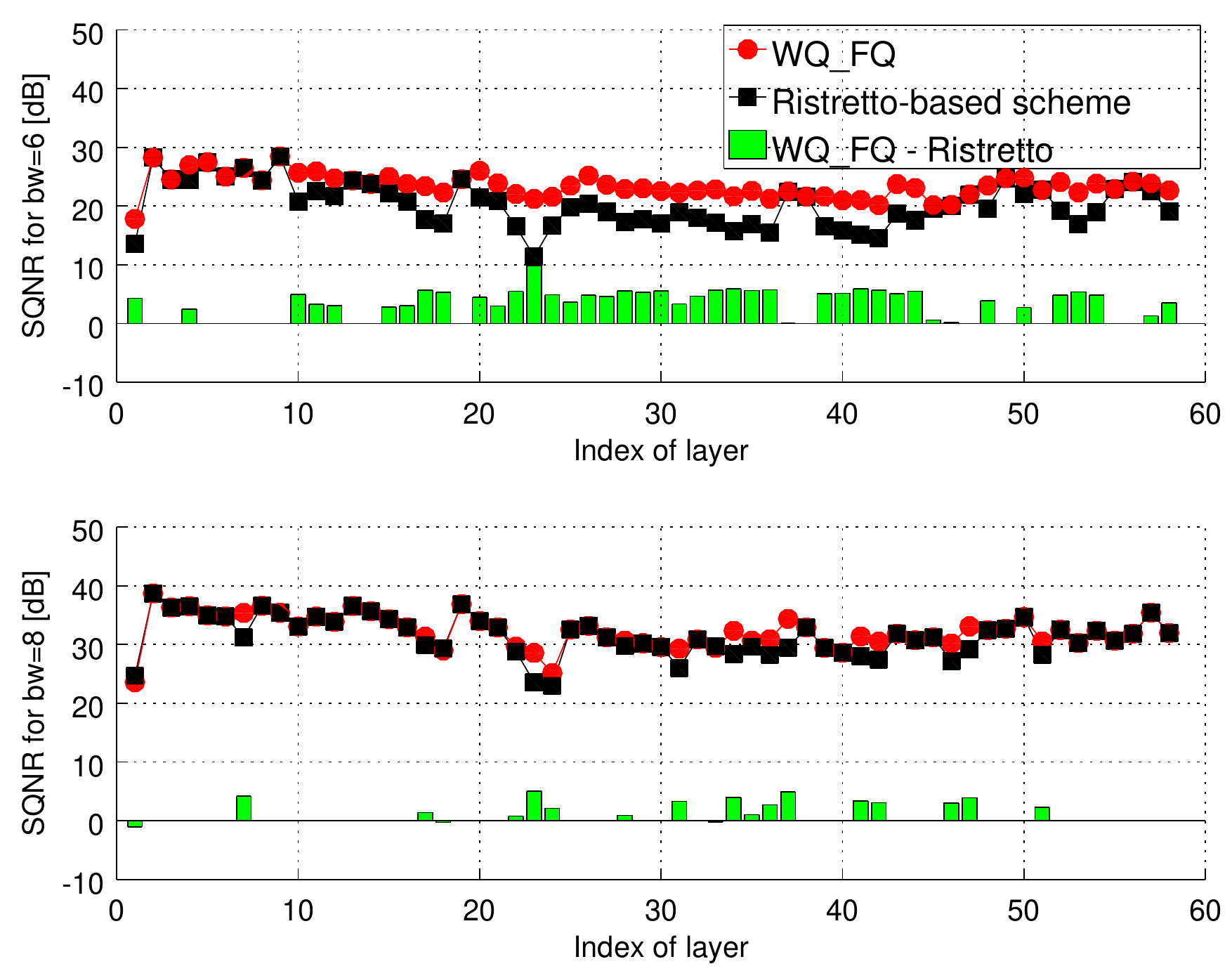}
\end{center}
   \caption{SQNRs for all layers after feature-map quantization.}
\label{fig:0013}
\end{figure}

\begin{figure}[tb]
\begin{center}
\includegraphics[width=0.9\linewidth]{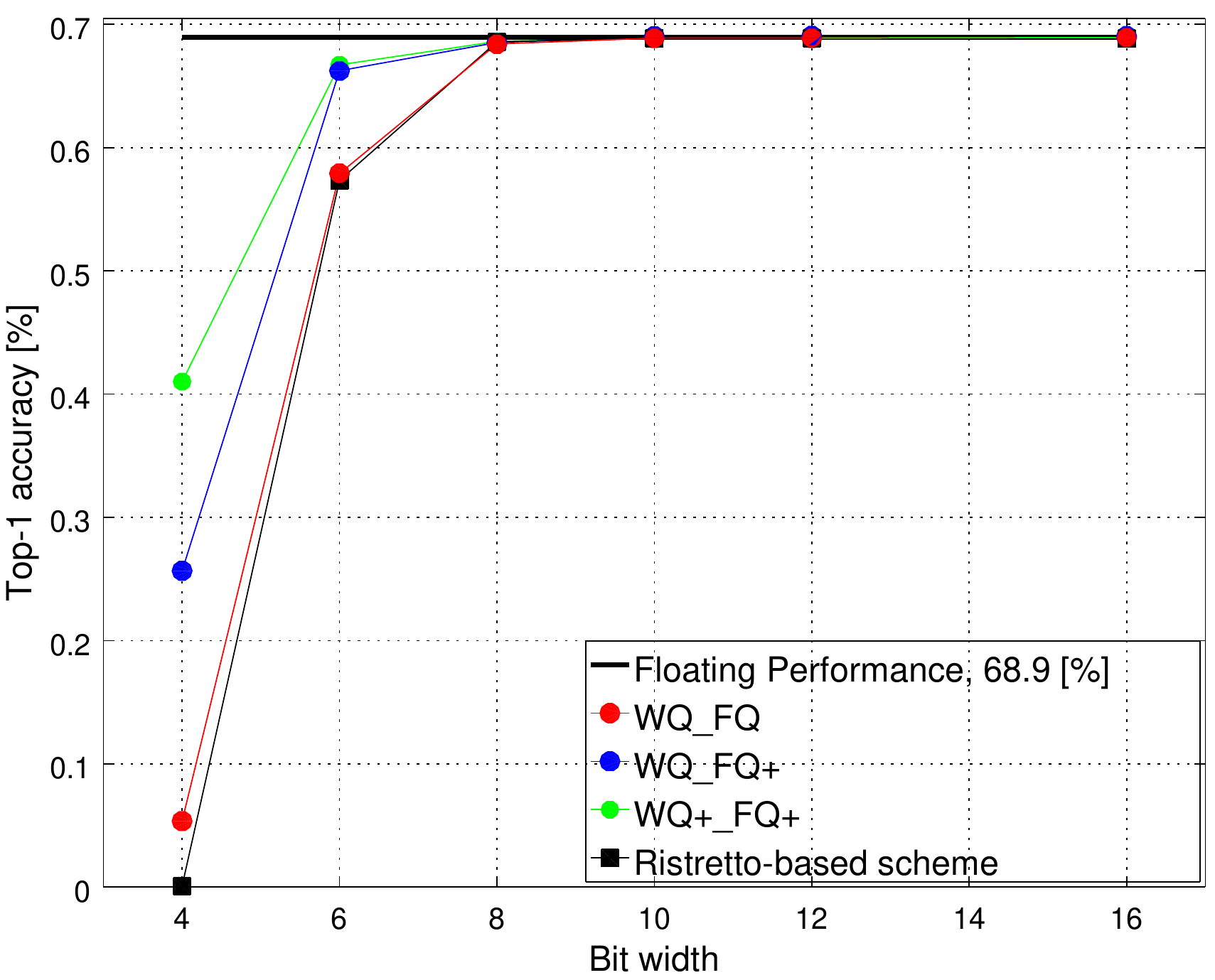}
\end{center}
   \caption{Top-1 accuracy of GoogLeNet after feature-map quantization.}
\label{fig:0014}
\end{figure}

In this section, we evaluate the performance in terms of SQNR and Top-1 accuracy after feature-map quantization steps. Figure \ref{fig:0013} shows SQNR for $WQ\_FQ$ and the Ristretto-based scheme, and SQNR difference between them for each layer. $WQ\_FQ$ has SQNR gain in many layers after feature-map quantization steps for $bw$ = 6. Figure \ref{fig:0014} shows Top-1 accuracy for five representations, that is, floating-point, $WQ\_FQ$, $WQ\_FQ^+$, $WQ^+\_FQ^+$, and a Ristretto-based scheme. All the schemes can almost achieve the floating-level performance when $bw \geq 8$. $WQ^+\_FQ^+$ has the performance gain about 40.9 \% for $bw$ = 4 and 9.3 \% for $bw$ = 6, over the Ristretto-based scheme. In Table \ref{tbl:0001}, Top-1 accuracies are summarized for all the schemes. $WQ\_FQ^+$ and $WQ^+\_FQ^+$ seem to be competitive as one of the powerful CNN quantization schemes when $bw \geq 6$.


\begin{table}
\begin{center}
\begin{tabular}{ c|c|c|c|c|c }
  \hline
  & \multicolumn{5}{c}{Top-1 accuracy [\%]} \\
  \hline
  \thead{Floating \\ Performance} & \multicolumn{5}{c}{68.9} \\
  \hline
  & 4 bit & 6 bit & 8 bit & 10 bit & 16 bit \\
  \hline
  \thead{Ristretto-based \\ scheme \\ (floating FM)} & 0.1 & 66.4 & 68.8 & 68.9 & 68.9 \\
  \thead{Ristretto-based \\ scheme } & 0.1 & 57.4 & 68.6 & 68.9 & 68.9 \\
  \hline
  \thead{$WQ$} & 38.8 & 67.4 & 68.6 & 68.9 & 68.9 \\
  \thead{$WQ^+$} & 54.8 & 68.0 & 68.8 & 68.9 & 68.9 \\
  \hline
  \thead{$WQ\_FQ$} & 5.4 & 57.9 & 68.4 & 68.9 & 68.9 \\
  \thead{$WQ\_FQ^+$}              & 25.7 & 66.2 & 68.5 & 69.1 & 69.1 \\
  \thead{$WQ^+\_FQ^+$}   & 41.0 & 66.7 & 68.6 & 69.1 & 69.1 \\
  \hline
\end{tabular}
\end{center}
\caption{Summary of Top-1 accuracies for several bit-widths}
\label{tbl:0001}
\end{table}

\subsection{Analysis on 4 bit quantization}
Our proposed scheme ($WQ^+ \_ FQ^+$ in Table \ref{tbl:0001}) has 27.9 \% performance degradation for 4 bit weight and 4 bit feature-map quantization. Therefore, it seems not acceptable to apply 4 bit quantization to GoogLeNet.
We try to find the dominant factor of performance degradation in the case. For the layers with a lower SQNR in weight quantization than a given threshold, 4 bit weights are replaced with 8 bit weights. We can find that Top-1 accuracy proportionally increases from 54.8 \% to 68.8 \% as the ratio of the number of layers of 8 bit quantized weights to the number of total layers increases, as shown in Table \ref{tbl:0002}. It is found that securing a high SQNR in weight quantization is requisite for achieving target performance, and the required SQNR should be at least 12 dB in order for performance loss to be less than 5 \% in the case of $WQ^+$.
Table \ref{tbl:0003} shows how Top-1 accuracy changes after feature-map quantization as the SQNR threshold of feature-maps increases. For the layers with a lower SQNR in feature-map quantization than a given threshold, 4 bit feature-maps are replaced with 8 bit feature-maps, while weight/bias quantization being kept to 8 bit. A very sharp drop in Top-1 accuracy is observed as the threshold changes 14dB to 13dB.

We find that the first convolutional layer ('$conv1/7\times7$') and the classifier ('$loss3/classifier$') are two critical layers to cause performance degradation in 4 bit feature-map quantization. From the analysis, we evaluate the alternative approach with 4 bit feature-map quantization and with 8 bit weight quantization for GoogLeNet. With $WQ^+ \_ FQ^+$, we finally achieve 65.5 \% Top-1 accuracy with only 4.4 \% degradation as seen in Table \ref{tbl:0004}.
\begin{table} [t]
\begin{center}
\begin{tabular}{ c|c|c|c|c|c }
  \hline
  & \multicolumn{5}{c}{Top-1 accuracy [\%]} \\
  \hline
  \thead{SQNR \\threshold [dB]} & $-\infty$ & 8 & 10 & 12 & $\infty$ \\
  \hline
  \thead{ratio [\%] \\ $\left( =\frac{\text{\# of layers of w8}}{\text{\# of all layers}}\right)$} & 0 & 21.9 & 42.2 & 60.9 & 100 \\
  \hline
  \thead{$WQ$} & 38.8 & 57.4 & 61.7 & 63.8 & 68.6 \\
  \thead{$WQ^+$} & 54.8 &  59.4 & 63.1 & 64.2 & 68.8 \\
  \hline
\end{tabular}
\end{center}
\caption{Top-1 accuracy after weight quantization according to SQNR thresholds of weights. Here w8 denotes weight quantization of 8 bits.}
\label{tbl:0002}
\end{table}

\begin{table} [t]
\begin{center}
\begin{tabular}{ c|c|c|c|c|c }
  \hline
  & \multicolumn{5}{c}{Top-1 accuracy under w8 [\%]} \\
  \hline
  \thead{SQNR \\threshold [dB]} & 13 & 14 & 15 & 16 & $\infty$ \\
  \hline
  \thead{ratio [\%] \\ $\left( =\frac{\text{\# of layers of fm8}}{\text{\# of all layers}}\right)$} & 10.6 & 37.9 & 68.2 & 81.8 & 100 \\
  \hline
  \thead{$WQ\_FQ$} & 5.1  & 65.7 & 67.5 & 68.0 & 68.4 \\
  \hline
\end{tabular}
\end{center}
\caption{Top-1 accuracy after feature-map quantization according to SQNR thresholds of feature-maps. Here w8 is assumed and fm8 means feature-map quantization of 8 bits.}
\label{tbl:0003}
\end{table}

\begin{table} [b]
\begin{center}
\begin{tabular}{ c|c|c }
  \hline
  & \multicolumn{2}{c}{Top-1 accuracy [\%]} \\
  \hline
  \thead{models} & \thead{(w8, fm4) \\w/ fm8 in classifier}  & \thead{(w8, fm4) \\w/ fm8 in classifier \\ and fm8 in the $1^{st}$ layer} \\
  \hline
  \thead{$WQ\_FQ$} & 61.0  & 64.0 \\
  \hline
  \thead{$WQ\_FQ^+$} & 62.9  & 64.9 \\
  \hline
  \thead{$WQ^+\_FQ^+$} & 63.0  & 65.5 (-4.4) \\
  \hline
\end{tabular}
\end{center}
\caption{Top-1 accuracy after (w8, fm4) quantization. Here (w8, fm4) denotes weight quantization of 8 bits and feature-map quantization of 4 bits.}
\label{tbl:0004}
\end{table}

\subsection{Comparison with the previous works}
Table \ref{tbl:0005} shows a comparison in GoogLeNet accuracy among previous quantization algorithms proposed for CNN. In the table 'Ours' denotes $WQ^+\_FQ^+$ which has the best performance among our schemes. For performance comparison, there are given two metrics, Top-1 accuracy (acc.) and performance loss (loss). Since there is not an identical floating-point accuracy for GoogLeNet as a reference, it seems better to compare each quantization algorithm in terms of fixed-point loss from floating-point. When $bw \geq 6$, our proposed algorithm has the lowest performance degradation compared to \cite{ref06:hubara2016quantized} and \cite{ref10:gysel2016hardware}. Our quantization algorithm achieves both high SQNR and high accuracy for $bw \geq 6$. Meanwhile, our algorithm is not the best when $bw \leq 4$. Quantization algorithms that are operated with their own architecture and with training from scratch have much better accuracy for $bw$ of small values. In actual commercial environments, however, datasets may not be sufficiently shared for quantization due to security and privacy issues, and hardware accelerators may not support a dedicated structure for a specific quantization algorithm, whereas our proposed quantization algorithm does not require any special structure as well as a dataset for training.

\begin{table} [t]
\begin{center}
\begin{tabular}{ c|c|c|c|c|c }
  \hline
  \multirow{2}{*}{} & float & \multicolumn{4}{c}{Top-1 accuracy [\%]} \\
  \cline{3-6}
  & (ref.) & & 4 bit & 6 bit & 8 bit \\
  \hline
  \multirow{2}{*}{Ours} & \multirow{2}{*}{68.9} & acc. & 41.0 & 66.7 & 68.6 \\
  \cline{3-6}
  & & loss & 27.9 & 2.2 & 0.3 \\


  \hline
  \multirow{2}{*}{Ristretto} & \multirow{2}{*}{68.9} & acc. & - & - & 66.6 \\
  \cline{3-6}
   & & loss & - & - & 2.3 \\

  \hline
  \multirow{2}{*}{QNN} & \multirow{2}{*}{71.6} & acc. & 66.5 & 66.4 & - \\
  \cline{3-6}
   & & loss & 5.1 & 5.2 & - \\

  \hline
\end{tabular}
\end{center}
\caption{Comparison with previous quantization algorithms in GoogLeNet. All accuracies have been evaluated from references: Ristretto \cite{ref10:gysel2016hardware} and QNN \cite{ref06:hubara2016quantized}}
\label{tbl:0005}
\end{table}

\subsection{Evaluation of various networks}
For generalization, we also evaluate top-1 accuracy in different networks such as AlexNet \cite{ref28:krizhevsky2012imagenet} and VGG-16 \cite{ref29:simonyan2014very}.
In Table \ref{tbl:0006}, a floating-level accuracy for both networks can be almost achieved by $WQ\_FQ^+$ of 8 bit and the performance degradation is less than 4\% by $WQ\_FQ^+$ of 6 bit.

\begin{table} [t]
\begin{center}
\begin{tabular}{ c|c|c|c|c|c }
  \hline
  \multirow{3}{*}{} & \multicolumn{5}{c}{Top-1 accuracy [\%]} \\
  \cline{2-6}
   & float & \multicolumn{2}{c |}{$WQ\_FQ$} & \multicolumn{2}{c}{$WQ\_FQ^+$} \\
  \cline{3-6}
   & (ref.)& 6 bit & 8 bit & 6 bit & 8 bit \\
  \hline
  {AlexNet} & 56.8 & 43.2 & 56.0 & 52.9 & 56.2 \\
  \hline
  {VGG-16} & 68.3 & 57.7 & 68.2 & 64.8 & 68.3 \\
  \hline
\end{tabular}
\end{center}
\caption{Top-1 accuracies for AlexNet and VGG-16, quantized by the proposed algorithms.}
\label{tbl:0006}
\end{table}


\section{Conclusion}
In this paper, we proposed an algorithm for weight, bias, and feature-map quantization in CNNs. The proposed algorithm has been designed to achieve the maximum SQNR in quantization steps and to enhance performance in tuning steps, separately. In quantization steps, we designed a simple SQNR-based quantization for weights/biases and a powerful GGD-based quantization for feature-maps. In tuning steps, moreover, we introduced the BFT algorithm to modify the fractional length to improve performance in a quantized network. By using the proposed algorithm, we achieved floating-level performance for AlexNet, VGG-16, and GoogLeNet with $bw=8$. For GoogLeNet, Top-1 accuracies of 66.7 \% and 65.5 \% have been achieved with (w6, fm6) and with (w8, fm4), respectively. In conclusion, the proposed algorithm is the powerful quantization approach to reduce power consumption and memory bandwidth for efficient implementation of hardware accelerators for inference.

\bibliography{quantization_dyunkim}

\begin{thebibliography}{22}
\providecommand{\natexlab}[1]{#1}
\providecommand{\url}[1]{\texttt{#1}}
\expandafter\ifx\csname urlstyle\endcsname\relax
  \providecommand{\doi}[1]{doi: #1}\else
  \providecommand{\doi}{doi: \begingroup \urlstyle{rm}\Url}\fi

\bibitem[Courbariaux et~al.(2014)Courbariaux, Bengio, and
  David]{ref04:courbariaux2014training}
Courbariaux, Matthieu, Bengio, Yoshua, and David, Jean-Pierre.
\newblock Training deep neural networks with low precision multiplications.
\newblock \emph{arXiv preprint arXiv:1412.7024}, 2014.

\bibitem[Courbariaux et~al.(2015)Courbariaux, Bengio, and
  David]{ref05:courbariaux2015binaryconnect}
Courbariaux, Matthieu, Bengio, Yoshua, and David, Jean-Pierre.
\newblock Binaryconnect: Training deep neural networks with binary weights
  during propagations.
\newblock In \emph{Advances in Neural Information Processing Systems}, pp.\
  3123--3131, 2015.

\bibitem[Deng et~al.(2009)Deng, Dong, Socher, Li, Li, and
  Fei-Fei]{ref25:deng2009imagenet}
Deng, Jia, Dong, Wei, Socher, Richard, Li, Li-Jia, Li, Kai, and Fei-Fei, Li.
\newblock Imagenet: A large-scale hierarchical image database.
\newblock In \emph{Computer Vision and Pattern Recognition, 2009. CVPR 2009.
  IEEE Conference on}, pp.\  248--255. IEEE, 2009.

\bibitem[Gom{\`e}s et~al.(2008)Gom{\`e}s, Combes, and
  Dussauchoy]{ref24:gomes2008parameter}
Gom{\`e}s, Oph{\'e}lie, Combes, Catherine, and Dussauchoy, Alain.
\newblock Parameter estimation of the generalized gamma distribution.
\newblock \emph{Mathematics and Computers in Simulation}, 79\penalty0
  (4):\penalty0 955--963, 2008.

\bibitem[Gupta et~al.(2015)Gupta, Agrawal, Gopalakrishnan, and
  Narayanan]{ref08:gupta2015deep}
Gupta, Suyog, Agrawal, Ankur, Gopalakrishnan, Kailash, and Narayanan, Pritish.
\newblock Deep learning with limited numerical precision.
\newblock In \emph{Proceedings of the 32nd International Conference on Machine
  Learning (ICML-15)}, pp.\  1737--1746, 2015.

\bibitem[Gysel et~al.(2016)Gysel, Motamedi, and
  Ghiasi]{ref10:gysel2016hardware}
Gysel, Philipp, Motamedi, Mohammad, and Ghiasi, Soheil.
\newblock Hardware-oriented approximation of convolutional neural networks.
\newblock \emph{arXiv preprint arXiv:1604.03168}, 2016.

\bibitem[Han et~al.(2015)Han, Mao, and Dally]{ref15:han2015deep}
Han, Song, Mao, Huizi, and Dally, William~J.
\newblock Deep compression: Compressing deep neural networks with pruning,
  trained quantization and huffman coding.
\newblock \emph{arXiv preprint arXiv:1510.00149}, 2015.

\bibitem[Han et~al.(2016)Han, Pool, Narang, Mao, Tang, Elsen, Catanzaro, Tran,
  and Dally]{ref16:han2016dsd}
Han, Song, Pool, Jeff, Narang, Sharan, Mao, Huizi, Tang, Shijian, Elsen, Erich,
  Catanzaro, Bryan, Tran, John, and Dally, William~J.
\newblock Dsd: Regularizing deep neural networks with dense-sparse-dense
  training flow.
\newblock \emph{arXiv preprint arXiv:1607.04381}, 2016.

\bibitem[Horowitz(2014)]{ref27:horowitz20141}
Horowitz, Mark.
\newblock 1.1 computing's energy problem (and what we can do about it).
\newblock In \emph{Solid-State Circuits Conference Digest of Technical Papers
  (ISSCC), 2014 IEEE International}, pp.\  10--14. IEEE, 2014.

\bibitem[Howard et~al.(2017)Howard, Zhu, Chen, Kalenichenko, Wang, Weyand,
  Andreetto, and Adam]{ref18:howard2017mobilenets}
Howard, Andrew~G, Zhu, Menglong, Chen, Bo, Kalenichenko, Dmitry, Wang, Weijun,
  Weyand, Tobias, Andreetto, Marco, and Adam, Hartwig.
\newblock Mobilenets: Efficient convolutional neural networks for mobile vision
  applications.
\newblock \emph{arXiv preprint arXiv:1704.04861}, 2017.

\bibitem[Hubara et~al.(2016)Hubara, Courbariaux, Soudry, El-Yaniv, and
  Bengio]{ref06:hubara2016quantized}
Hubara, Itay, Courbariaux, Matthieu, Soudry, Daniel, El-Yaniv, Ran, and Bengio,
  Yoshua.
\newblock Quantized neural networks: Training neural networks with low
  precision weights and activations.
\newblock \emph{arXiv preprint arXiv:1609.07061}, 2016.

\bibitem[Hui \& Neuhoff(2001)Hui and Neuhoff]{ref19:hui2001asymptotic}
Hui, Dennis and Neuhoff, David~L.
\newblock Asymptotic analysis of optimal fixed-rate uniform scalar
  quantization.
\newblock \emph{IEEE Transactions on Information Theory}, 47\penalty0
  (3):\penalty0 957--977, 2001.

\bibitem[Iandola et~al.(2016)Iandola, Han, Moskewicz, Ashraf, Dally, and
  Keutzer]{ref17:iandola2016squeezenet}
Iandola, Forrest~N, Han, Song, Moskewicz, Matthew~W, Ashraf, Khalid, Dally,
  William~J, and Keutzer, Kurt.
\newblock Squeezenet: Alexnet-level accuracy with 50x fewer parameters and< 0.5
  mb model size.
\newblock \emph{arXiv preprint arXiv:1602.07360}, 2016.

\bibitem[Jia et~al.(2014)Jia, Shelhamer, Donahue, Karayev, Long, Girshick,
  Guadarrama, and Darrell]{ref26:jia2014caffe}
Jia, Yangqing, Shelhamer, Evan, Donahue, Jeff, Karayev, Sergey, Long, Jonathan,
  Girshick, Ross, Guadarrama, Sergio, and Darrell, Trevor.
\newblock Caffe: Convolutional architecture for fast feature embedding.
\newblock In \emph{Proceedings of the 22nd ACM international conference on
  Multimedia}, pp.\  675--678. ACM, 2014.

\bibitem[Judd et~al.(2015)Judd, Albericio, Hetherington, Aamodt, Jerger,
  Urtasun, and Moshovos]{ref11:judd2015reduced}
Judd, Patrick, Albericio, Jorge, Hetherington, Tayler, Aamodt, Tor, Jerger,
  Natalie~Enright, Urtasun, Raquel, and Moshovos, Andreas.
\newblock Reduced-precision strategies for bounded memory in deep neural nets.
\newblock \emph{arXiv preprint arXiv:1511.05236}, 2015.

\bibitem[Krizhevsky et~al.(2012)Krizhevsky, Sutskever, and
  Hinton]{ref28:krizhevsky2012imagenet}
Krizhevsky, Alex, Sutskever, Ilya, and Hinton, Geoffrey~E.
\newblock Imagenet classification with deep convolutional neural networks.
\newblock In \emph{Advances in neural information processing systems}, pp.\
  1097--1105, 2012.

\bibitem[Lin et~al.(2016)Lin, Talathi, and Annapureddy]{ref14:lin2016fixed}
Lin, Darryl, Talathi, Sachin, and Annapureddy, Sreekanth.
\newblock Fixed point quantization of deep convolutional networks.
\newblock In \emph{International Conference on Machine Learning}, pp.\
  2849--2858, 2016.

\bibitem[Long et~al.(2015)Long, Shelhamer, and Darrell]{ref03:long2015fully}
Long, Jonathan, Shelhamer, Evan, and Darrell, Trevor.
\newblock Fully convolutional networks for semantic segmentation.
\newblock In \emph{Proceedings of the IEEE Conference on Computer Vision and
  Pattern Recognition}, pp.\  3431--3440, 2015.

\bibitem[Ren et~al.(2015)Ren, He, Girshick, and Sun]{ref02:ren2015faster}
Ren, Shaoqing, He, Kaiming, Girshick, Ross, and Sun, Jian.
\newblock Faster r-cnn: Towards real-time object detection with region proposal
  networks.
\newblock In \emph{Advances in neural information processing systems}, pp.\
  91--99, 2015.

\bibitem[Simonyan \& Zisserman(2014)Simonyan and
  Zisserman]{ref29:simonyan2014very}
Simonyan, Karen and Zisserman, Andrew.
\newblock Very deep convolutional networks for large-scale image recognition.
\newblock \emph{arXiv preprint arXiv:1409.1556}, 2014.

\bibitem[Stacy \& Mihram(1965)Stacy and Mihram]{ref23:stacy1965parameter}
Stacy, E~Webb and Mihram, G~Arthur.
\newblock Parameter estimation for a generalized gamma distribution.
\newblock \emph{Technometrics}, 7\penalty0 (3):\penalty0 349--358, 1965.

\bibitem[Szegedy et~al.(2015)Szegedy, Liu, Jia, Sermanet, Reed, Anguelov,
  Erhan, Vanhoucke, and Rabinovich]{ref01:szegedy2015going}
Szegedy, Christian, Liu, Wei, Jia, Yangqing, Sermanet, Pierre, Reed, Scott,
  Anguelov, Dragomir, Erhan, Dumitru, Vanhoucke, Vincent, and Rabinovich,
  Andrew.
\newblock Going deeper with convolutions.
\newblock In \emph{Proceedings of the IEEE conference on computer vision and
  pattern recognition}, pp.\  1--9, 2015.

\end{thebibliography}
\bibliographystyle{icml2018}

\end{document}